\documentclass[twocolumn]{article}
\usepackage[dvipdfmx]{graphicx}
\usepackage{arxiv}
\usepackage{authblk}

\usepackage{multirow}%
\usepackage{amsmath,amssymb,amsfonts,amsthm}
\usepackage{mathrsfs}%
\usepackage[utf8]{inputenc}
\usepackage{hyperref}
\usepackage{enumitem}

\usepackage{xcolor}%
\usepackage{textcomp}%
\usepackage{manyfoot}%
\usepackage{booktabs}%
\usepackage{algorithm}%
\usepackage{algorithmicx}%
\usepackage{algpseudocode}%
\usepackage{listings}%

\usepackage{multirow}
\usepackage{subfig}
\usepackage{array}
\usepackage{booktabs}
\usepackage{threeparttable}
\usepackage{colortbl}
\usepackage{siunitx}
\usepackage{tikz}
\usetikzlibrary{positioning,arrows.meta,fit,calc,shapes.geometric}

\theoremstyle{thmstyleone}%
\theoremstyle{thmstyletwo}%

\theoremstyle{thmstylethree}%

\newcolumntype{P}[1]{>{\centering\arraybackslash}m{#1}}

\begin{document}

\newcommand{\myPaperShortTitle}{Three-in-One World Model}
\newcommand{\myPaperTitle}{Three-in-One World Model: Energy-Based Consistency, Prediction, and Counterfactual Inference for Marketing Intervention}
\title{\myPaperTitle}
\date{}

%\author{Junichiro Niimi}

\renewcommand\Authfont{\bfseries}
\setlength{\affilsep}{0em}
% box is needed for correct spacing with authblk
\newbox{\orcid}\sbox{\orcid}{\includegraphics[scale=0.06]{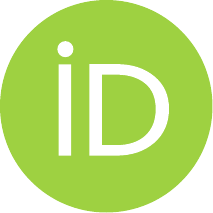}} 
\author[1]{%
	\href{https://orcid.org/0000-0002-4618-6272}{\usebox{\orcid}\hspace{1mm}
	Junichiro Niimi\thanks{\texttt{jniimi@meijo-u.ac.jp}}
	}}
\affil[1]{Meijo University}

\renewcommand{\shorttitle}{\myPaperShortTitle}
\newcommand{\FOne}{macro\text{-}F_1}
\newcommand{\RMSE}{R\hspace{-0.05em}M\hspace{-0.1em}S\hspace{-0.07em}E}

%\maketitle

\twocolumn[
	\begin{@twocolumnfalse}
		\maketitle
\vspace{-3em}
\begin{abstract}
Marketing decisions reflect the interaction of latent consumer heterogeneity, time-varying internal states, and explicit interventions, a structure that current prediction- and language-oriented models do not capture in a unified manner. We propose a Three-in-One world-model architecture in which a Deep Boltzmann Machine (DBM) learns a frozen belief representation from demographics, time, and lagged actions and outcomes, with lightweight task-specific adapters attached on top. The same belief supports three tasks within a single framework: (i) energy-based consistency evaluation through the DBM's free energy, (ii) outcome prediction through adapters, and (iii) counterfactual inference by holding the belief fixed and varying only the action input given to the adapter. Using a controlled simulation in which the latent price sensitivity, promotion responsiveness, and base preference of each consumer are known, we show that the adapters match a strong MLP baseline on visit- and purchase-AUC while recovering heterogeneous treatment effects substantially better than S-, T-, X-, and DR-learner meta-learners and a Causal Forest baseline built on the same raw features, with the largest gap on a confounded price-promotion intervention. Complementing this, free-energy clamps systematically penalize counterfactual purchase trajectories that lack prior promotional exposure, and the penalty itself depends on the latent base preference in the expected direction. These results indicate that DBM beliefs disentangle latent traits in a form that survives counterfactual queries, providing an integrated world-model substrate for marketing intervention.
\end{abstract}
\vspace{0.5em}
\vspace{2em}
	\end{@twocolumnfalse}
]

\setcounter{footnote}{0}
\section{Introduction}
\subsection{Background} % Section 1.1

Large language models (LLMs) have achieved remarkable progress in recent years. Since the introduction of the Transformer architecture \cite{transformer}, the field has witnessed rapid scaling of autoregressive language models, such as generative pretrained transformer (GPT) \cite{gpt1,gpt2,gpt3,gpt4}, with performance improving predictably as a function of model and data size \cite{openai-scaling-laws}. These models now demonstrate impressive capabilities across a wide range of tasks, yet their success relies fundamentally on next-token prediction over text sequences, a paradigm that lacks explicit mechanisms for causal reasoning or internal world representation \cite{lecun2022}.

In contrast to such token-level prediction, world models {\it i.e.,} systems that learn an internal representation of environment dynamics to enable prediction, planning, and reasoning, have attracted growing attention as a complementary paradigm \cite{ha2018,lecun2022}. Recent theoretical proposals \cite{lecun2022} and empirical advances \cite{ha2018,i-jepa,v-jepa} have positioned world models as a promising path beyond the limitations of autoregressive generation. This view is echoed in broader discourse: for example, Yann LeCun has described autoregressive LLMs as ``doomed'' due to their inability to model causality\footnote{Joint Mathematics Meetings, January 2025.}, and Demis Hassabis has emphasized that artificial general intelligence (AGI) requires world models capable of understanding physical reality\footnote{Google DeepMind Podcast, December 2025.}.

To date, world models have been explored predominantly in the context of physical environments ({\it e.g.,} robotic control \cite{ha2018} and video generation \cite{v-jepa,image_world_model}) where the goal is to simulate and anticipate interactions with the real world \cite{worldmodel_survey_ding}. Yet the core premise of world models, learning the latent dynamics that govern sequential observations, applies equally to behavioral domains. In consumer behavior, for instance, purchase decisions emerge from the interaction of latent individual traits, time-varying states, and marketing interventions, a dynamic process that demands the same kind of internal representation that world models provide for physical systems.

This paper therefore proposes a world-model architecture with multiple task-specific adapters that captures the structure of consumer behavior and supports prediction, consistency evaluation, and counterfactual inference within a single framework. We evaluate the proposed model using simulation data, where ground-truth counterfactual outcomes are available for rigorous assessment of causal effect recovery. The remainder of the paper is organized as follows. Section 2  reviews the related study to formulate the research gaps. Section 3 introduces the architecture and algorithm of the proposed model. Section 4 conducts the three simulation-based experiments to demonstrate the performance of the proposed model. Finally, Section 5 summarizes the findings, implications, and limitations of the study.

\subsection{Contributions} % Section 1.2
This study makes four contributions.

\paragraph{Marketing application of world models.} We bring the world-model paradigm \cite{ha2018,lecun2022} into the marketing domain by treating the consumer-and-environment dynamics---demographics, time, lagged interventions, and lagged outcomes---as the system whose latent structure is to be learned. The resulting model captures time-varying consumer states using only lagged features in the visible layer, without recurrent connections, while preserving the asymmetry between a heavy world model and lightweight downstream modules that characterizes \cite{ha2018}.

\paragraph{Energy-based consistency evaluation.} By adopting a DBM as the world model, the proposed framework can score the consistency of an input pattern through its free energy: low free energy indicates that the configuration of demographics, history, and observed behavior is plausible under the learned distribution, whereas high free energy flags an implausible scenario. Unlike the static co-occurrence model of \cite{niimi_boltzmanngpt}, the consistency check here applies to time-extended trajectories that include lagged actions and outcomes.

\paragraph{Prediction via belief representation.} The belief vector formed by concatenating activations of all hidden layers serves as a frozen, task-agnostic substrate on which compact MLP adapters are trained for each downstream task. Empirically, this two-stage design matches the test-set AUC of an MLP trained end-to-end on the raw features, demonstrating that the dimensionality reduction performed by the DBM does not sacrifice predictive power.

\paragraph{Counterfactual inference.} Because the current-period intervention vector is supplied only to the adapter and not to the DBM, counterfactual outcomes can be obtained by fixing the belief and toggling the action input. This yields individual-level CATE estimates that, under the controlled simulation studied here, recover the true heterogeneity in price sensitivity and promotion responsiveness more accurately than S-, T-, X-, and DR-learner meta-learners and a Causal Forest baseline that operate on the same raw features.

\section{Related Study}
\subsection{Large Language Models}

Building on the background in Section~1, we here review LLMs with a focus on their real-world applications and their relationship to world models. Through next-token prediction on massive corpora, LLMs acquire broad linguistic and reasoning capabilities that transfer to a wide range of downstream tasks \cite{gpt3,llm_sentiment_review,llm_finance_review,llm_education_review}.

In the real-world application, particularly in the field of marketing, LLMs have been applied primarily to text-centric tasks such as sentiment analysis \cite{llm_sentiment_review}, review generation \cite{generate_review_llm1,llm_marketing_review}, and consumer simulation \cite{llm_persona,tseng_llm_persona}. While these applications exploit the linguistic fluency of LLMs, they do not address the structural modeling of consumer dynamics---the latent heterogeneity, time-varying states, and intervention effects---that is the focus of the present study.

A central open question is whether LLMs implicitly acquire world models through language modeling alone. Some evidence suggests that LLMs can develop internal representations of environment states. A representative line of work on board games began with Othello-GPT \cite{othellogpt}, demonstrating that world representations emerge in a synthetic task without any prior knowledge about the game rules \cite{nanda2023,karvonen2024,spies2025}.

Other studies have focused on LLM-driven agent simulations, demonstrating emergent social behaviors that resemble aspects of world modeling \cite{simulacra,opencity}. However, without additional architectural constraints, such representations remain implicit, entangled with linguistic knowledge, and do not readily support causal or counterfactual reasoning \cite{lecun2022}.

More fundamentally, a growing body of criticism questions whether fluent generation equates to genuine understanding. Bender and Koller \cite{bender_acl2020} argue that a system trained solely on linguistic form cannot acquire meaning, because the mapping from form to communicative intent is underdetermined by text alone. Bender et al. \cite{bender2021parrots} further characterize LLMs as ``stochastic parrots'' that reproduce statistical patterns in their training data without grounded comprehension. These critiques suggest that next-token prediction, however scaled, may be insufficient for building models that truly represent how the world works---motivating the development of explicit world models.

\subsection{World Models}
\subsubsection{Two Major Trends in World Models}
World models have gained renewed attention as a complementary paradigm to LLMs \cite{worldmodel_survey_ding,worldmodel_survey_mai}. The field encompasses two major streams: reinforcement-learning-based approaches \cite{ha2018,schmidhuber2015} and representation-learning approaches \cite{lecun2022,v-jepa,i-jepa}. In this study, we draw on ideas from both streams.

First, Ha and Schmidhuber \cite{ha2018} introduced an RNN-based world model that learns environmental dynamics through state prediction, comprising Vision ($V$), Memory ($M$), and Controller ($C$) modules. $M$ predicts the latent state at time $t+1$ from the current latent state at time $t$ and action, while $C$ selects actions based on the resulting representation. A key design principle is the asymmetry between components: $C$ (867 parameters) is orders of magnitude smaller than $V$ (4.3M) or $M$ (0.4M), and $C$ operates only on $z$, an abstract representation produced by $V$. This demonstrates that if a world model captures sufficient structure, the downstream controller need not be large.

Second, the Joint Embedding Predictive Architecture (JEPA) line of work \cite{i-jepa,v-jepa} operationalizes LeCun's vision \cite{lecun2022} by learning world models in an abstract embedding space, demonstrating that meaningful structure can be captured without reconstructing pixel-level details as in generative approaches.
Central to this framework is the use of Energy-Based Models (EBMs) \cite{ebm_tutorial}, in which an energy function distinguishes plausible from implausible configurations, as a foundation for world models: low energy indicates a plausible pattern, while high energy signals an implausible one. This property is particularly attractive for world models, because the energy landscape provides a built-in mechanism for evaluating input consistency---a capability that has been exploited for anomaly detection \cite{ebm_anomaly} and, more recently, for cognitively inspired world models \cite{energy_based_transformer}.

\subsubsection{Deep Boltzmann Machine}
The Deep Boltzmann Machine (DBM) \cite{dbm} is a representative EBM that stacks multiple layers of Restricted Boltzmann Machines (RBMs) \cite{smolensky_rbm,hinton_rbm}, each an undirected bipartite model whose intra-layer independence enables efficient inference. The DBM is trained via greedy layerwise pretraining of individual RBMs followed by joint fine-tuning across all layers \cite{bm_training}. As a generative model, the DBM assigns an energy to every configuration of visible and hidden units, and its free energy (FE) provides a principled score of how well an input pattern conforms to the learned distribution.

Niimi \cite{niimi_boltzmanngpt} adopted a DBM as a world model and combined it with an adapter and GPT-2 \cite{gpt2} to build a consumer review generation model. By separating the world model from the language model, the study demonstrated that (i) energy-based coherence can be evaluated by the DBM alone, (ii) conditioning through the world model enables a small language model to generate fluent and semantically coherent text, and (iii) interventional generation is possible by clamping the input vector of the DBM. However, that model was limited to static co-occurrence among variables and could not capture time-varying dynamics within individual consumers.

Several studies have extended RBMs to handle time-series data \cite{crbm,fcrbm,rtrbm,rnnrbm,srtrbm}. These extensions were designed primarily for sequential generation tasks such as music composition and motion capture, where modeling the conditional distribution of the next observation given the history is the natural objective. The Conditional RBM (CRBM) \cite{crbm} achieves this by using past observations to modulate biases at each time step, while models with recurrent structures, such as Recurrent Temporal RBM (RT-RBM) \cite{rtrbm} and RNN-RBM \cite{rnnrbm}, introduce hidden-to-hidden connections trained via backpropagation through time (BPTT) \cite{mlp,bptt,bptt2}.

Our design goals differ. For counterfactual inference, we require a world model whose parameters represent the stable structure of the environment, with temporal variation captured in the agent's internal state (belief) rather than in time-varying model parameters. We therefore adopt a standard DBM and incorporate temporal information through lagged features in the visible layer. This simpler design also enables the use of a multi-layer hierarchy, which can in principle provide a richer, hierarchically summarized belief representation by concatenating the activations of all hidden layers---an option not exploited by the temporal RBM extensions above, which typically operate with a single hidden layer.

\subsection{Causal Inference}

Causal inference aims to estimate the effect of an intervention from observational or experimental data. Two foundational frameworks underpin this field: the potential outcomes framework \cite{rubin2005causal}, which defines causal effects through the contrast of counterfactual outcomes $Y_i(1)$ and $Y_i(0)$ for each unit $i$, and the structural causal model of Pearl \cite{causality}, which formalizes interventions via the $do$-operator. A central challenge is that only one potential outcome is ever observed for a given unit---the fundamental problem of causal inference.

When treatment effects vary across individuals, the quantity of interest is the Conditional Average Treatment Effect (CATE): $\tau(x) = E[Y(1) - Y(0) \mid X = x]$. K\"{u}nzel et al.\ \cite{metalearners} systematize CATE estimation under the meta-learner framework. The S-learner fits a single model with treatment as a feature and estimates CATE as $\hat{\mu}(x, 1) - \hat{\mu}(x, 0)$; the T-learner trains separate models for treated and control groups. The X-learner \cite{metalearners} extends the T-learner by imputing individual treatment effects on each arm and combining them through a propensity-weighted average, which improves performance under unbalanced treatment assignment. The DR-learner \cite{drlearner} regresses a doubly robust pseudo-outcome---constructed from outcome-model and propensity-score nuisance estimates with cross-fitting---on covariates, providing optimality guarantees under flexible nuisance estimation. Wager and Athey \cite{causalforest} propose the Causal Forest, a nonparametric approach based on random forests that partitions the covariate space to estimate heterogeneous effects directly. These methods serve as baselines in our experiments.

In marketing, CATE estimation is operationalized as uplift modeling ({\it i.e.,} the task of identifying which customers will respond positively to a specific intervention such as a promotion or advertisement) \cite{radcliffe2007upliftmodeling,diemert2018upliftmodeling,upliftmodeling_decisiontree}. Unlike conventional predictive models that estimate the outcome level, uplift models estimate the incremental effect of treatment, enabling targeted intervention strategies. Our approach differs from existing uplift methods in that CATE is derived from the counterfactual predictions of a world model: by holding the belief representation fixed and varying only the action input, the adapter produces paired predictions whose difference yields an individual-level treatment effect estimate.

\subsection{Research Gaps}
The related studies reviewed above reveal several open issues that the present work aims to address.

\paragraph{World model for consumer behavior remains static.} Existing work on DBM-based world models for marketing \cite{niimi_boltzmanngpt} operates solely on static, non-time-varying variables. As a result, the model cannot represent how consumer states evolve over time in response to past actions and outcomes.

\paragraph{Coherence evaluation is limited to static co-occurrence.} Because the underlying world model is static, energy-based coherence evaluation is restricted to assessing co-occurrence among consumer demographics, product attributes, and behavioral logs at a single point in time. It cannot evaluate the consistency of temporal scenarios---for example, whether a sequence of purchase behaviors is plausible given a consumer's history and the interventions they received.

\paragraph{Downstream tasks lack modularity.} In most multi-task learning approaches, accommodating a new downstream task requires updating the shared model weights, risking catastrophic forgetting of previously learned tasks. A world model whose internal representation is frozen after training would allow new task-specific adapters to be added without disturbing the learned representation, but this modular design has not been explored in the marketing context.

\paragraph{Standard predictive models are insufficient for CATE recovery.} Conventional predictive models such as MLPs, when used as S-learners, estimate outcomes rather than treatment effects directly. Without an explicit mechanism for counterfactual comparison---such as holding a latent representation fixed while varying the treatment input---these models struggle to recover heterogeneous causal effects, particularly when treatment assignment is correlated with latent consumer traits.

For these reasons, we adopt a standard DBM for the world model and incorporate temporal information simply through lagged features in the visible layer, avoiding the complexity and instability of recurrent extensions.

\section{Proposed Model}\label{sec:model}

\begin{figure*}[t]
\centering
\includegraphics[width=\linewidth]{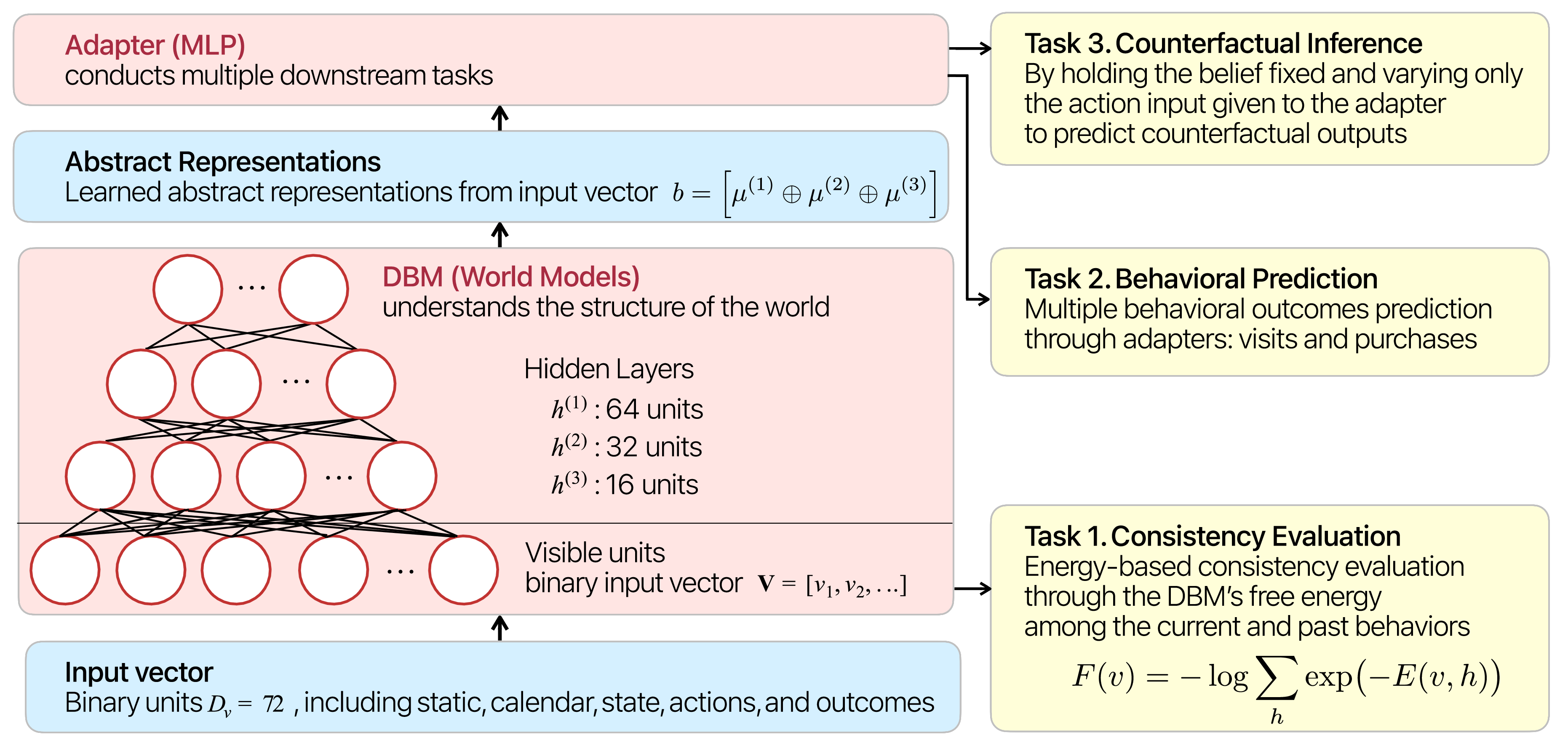}
\caption{\textbf{Three-in-One world-model architecture.} The visible vector $v$ enters a frozen Bernoulli--Bernoulli DBM with hidden layers of size $64$--$32$--$16$. The belief $b$, formed by concatenating mean-field activations of all three hidden layers, supports three task interfaces on a single shared substrate: \textbf{(i)} the free energy $F(v)$ scores the consistency of the input trajectory; \textbf{(ii)} a small task-specific adapter $g_k$ takes $[b;z]$ and outputs the predictive probability $\hat p_k$; \textbf{(iii)} counterfactual outcomes are obtained by holding $b$ fixed and varying only $z$, since the action vector never enters the DBM.}
\label{fig:arch}
\end{figure*}

We propose a world-model architecture that combines a Deep Boltzmann Machine (DBM) with multiple task-specific adapters, supporting three tasks within a single framework: (i) energy-based consistency evaluation of consumer trajectories, (ii) prediction of downstream outcomes such as visit and purchase, and (iii) counterfactual inference of intervention effects. The world model and the adapters are trained in two stages, and the world model's parameters are frozen before any adapter is fit, so that adding a new downstream task requires training only a new adapter. Figure~\ref{fig:arch} summarizes the architecture.

\subsection{Visible Layer and Lagged Features}\label{sec:visible}

For each consumer $i$ and time index $t$, we construct a binary visible vector $v_{i,t} \in \{0,1\}^{D_v}$ by concatenating five blocks: (a) static consumer attributes (store, loyalty bin, age decade, income quintile), (b) calendar features (day-of-week, month), (c) time-varying state summaries derived from the consumer's history (recent visit/purchase recency dummies and cumulative visit/purchase count thresholds), (d) lagged action indicators for the past $ws$ periods (\texttt{coupon}, \texttt{campaign}, \texttt{push}), and (e) lagged outcome indicators (\texttt{visit}, \texttt{purchase}) over the same window. All non-binary attributes are one-hot or threshold-encoded. With $ws=4$ the total dimension is $D_v=72$; Table~\ref{tab:visible} gives the per-block breakdown.

\begin{table*}[t]
\centering
\small
\caption{Visible-layer composition with $ws=4$ ($D_v=72$). Variable names match the columns in the simulated panel data.}
\label{tab:visible}
\setlength{\tabcolsep}{4pt}
\renewcommand{\arraystretch}{1.15}
\begin{tabular}{
@{}
llr
p{1.0\columnwidth}
@{}
}
\toprule
Block & Variable & Dim & Description \\
\midrule
\multirow{6}{*}{\textbf{Static}}
 & \texttt{store\_\{0..9\}} & 10 & One-hot of the consumer's home store ($10$ stores) \\
 & \texttt{loyalty} & 1 & $1$ if the consumer is enrolled in the loyalty program \\
 & \texttt{age\_\{20..60\}s} & 5 & Decade one-hot of age \\
 & \texttt{income\_q\{1..5\}} & 5 & Quintile one-hot of annual income \\
 & \texttt{month\_\{1..12\}} & 12 & One-hot of calendar month \\
 & \texttt{dow\_\{0..6\}} & 7 & One-hot of day-of-week (Monday $=0$) \\
\midrule
\multirow{9}{*}{\textbf{Time-varying}}
 & \texttt{visited\_within\_\{7,14,30\}d} & 3 & $1$ if any visit occurred within the last $7/14/30$ days \\
 & \texttt{purchased\_within\_\{7,14,30\}d} & 3 & $1$ if any purchase occurred within the last $7/14/30$ days \\
 & \texttt{cum\_visits\_\{5,10,30\}plus} & 3 & $1$ if cumulative visits to date $\ge 5/10/30$ \\
 & \texttt{cum\_purchases\_\{5,10,30\}plus} & 3 & $1$ if cumulative purchases to date $\ge 5/10/30$ \\
 & \texttt{coupon\_lag\{1..4\}} & 4 & Coupon issued $1$--$4$ days ago \\
 & \texttt{campaign\_lag\{1..4\}} & 4 & Store-level campaign active $1$--$4$ days ago \\
 & \texttt{push\_lag\{1..4\}} & 4 & Push notification sent $1$--$4$ days ago \\
 & \texttt{visit\_lag\{1..4\}} & 4 & Visit observed $1$--$4$ days ago \\
 & \texttt{purchase\_lag\{1..4\}} & 4 & Purchase observed $1$--$4$ days ago \\
\midrule
\multicolumn{2}{@{}l}{\textbf{Total}} & \textbf{72} & \\
\bottomrule
\end{tabular}
\end{table*}

The current-period action vector $z_{i,t} \in \{0,1\}^{D_z}$ (\texttt{sale1}, \texttt{sale2}, \texttt{campaign}, \texttt{coupon}, \texttt{push}) is deliberately \emph{not} placed in the visible layer of the DBM. It is supplied directly to the adapter (Section~\ref{sec:adapter}). This separation has two consequences. First, the DBM learns a representation of the consumer state that is independent of the current intervention, so the same belief can be reused across counterfactual queries. Second, the adapter is forced to learn the interaction between the consumer state and the action, which is precisely the quantity targeted by counterfactual prediction.

\subsection{World Model: Deep Boltzmann Machine}\label{sec:dbm}

The world model is a Bernoulli--Bernoulli DBM with $L$ stacked hidden layers $h^{(1)}, \dots, h^{(L)}$. Layer sizes are $64$--$32$--$16$ in our experiments. The joint energy is
\begin{align}\label{eq:joint_energy}
E(v, h) ={} & -v^{\top} W^{(1)} h^{(1)} - \sum_{\ell=1}^{L-1} h^{(\ell)\top} W^{(\ell+1)} h^{(\ell+1)} \notag\\
            & \quad - b_v^{\top} v - \sum_{\ell=1}^{L} b_{\ell}^{\top} h^{(\ell)},
\end{align}
where $W^{(\ell)}$ are inter-layer weight matrices and $b_v, b_{\ell}$ are biases. The free energy of a visible configuration,
\begin{equation}
F(v) = -\log \sum_{h} \exp\bigl(-E(v,h)\bigr),\label{eq:freeenergy}
\end{equation}
is the principal scalar score used for consistency evaluation.

Training follows the standard two-stage procedure for DBMs \cite{dbm,bm_training}: each pair of adjacent layers is greedily pretrained as an RBM, after which the full network is fine-tuned jointly using persistent contrastive divergence with mean-field inference for the data-dependent term. We use the Adam optimizer \cite{adam} for both phases.

\subsection{Belief Representation}\label{sec:belief}

After training, the DBM is frozen and used as a feature extractor. For each visible vector $v_{i,t}$ we run mean-field inference to obtain layerwise activations $\mu^{(\ell)}_{i,t} \in [0,1]^{d_\ell}$ and define the belief vector as the concatenation of all hidden activations,
\begin{equation}
b_{i,t} = \bigl[\mu^{(1)}_{i,t};\, \mu^{(2)}_{i,t};\, \mu^{(3)}_{i,t}\bigr] \in [0,1]^{d_1 + d_2 + d_3}.
\end{equation}
Including all layers rather than only the topmost layer preserves information at multiple levels of abstraction, allowing each adapter to draw on whichever scale is most informative for its task. With layer sizes $64$--$32$--$16$, the belief has dimension $112$.

\subsection{Adapters}\label{sec:adapter}

For each downstream task $k$ (e.g., visit prediction, purchase prediction) we train a task-specific adapter $g_k$, instantiated as a small MLP, that takes as input the concatenation of the frozen belief and the current-period action vector,
\begin{equation}
\hat{p}_{k,i,t} = \sigma\bigl(g_k\bigl([b_{i,t};\, z_{i,t}]\bigr)\bigr).
\end{equation}
The world-model parameters are not updated during adapter training, so adding a new task does not perturb the belief representation or any previously trained adapter, avoiding catastrophic forgetting across tasks.

\subsection{Three Tasks on a Shared Belief}\label{sec:tasks}

\paragraph{Consistency evaluation.} The free energy in \eqref{eq:freeenergy} is computed analytically for the Bernoulli--Bernoulli DBM and serves as a real-valued plausibility score for any visible configuration $v_{i,t}$. Because $v_{i,t}$ contains lagged actions and outcomes, this score evaluates the consistency of \emph{trajectories} rather than static co-occurrence, addressing a limitation of \cite{niimi_boltzmanngpt}.

\paragraph{Prediction.} The adapter outputs the predictive probability of each downstream outcome given the current consumer state and the current intervention. Because the belief is frozen, the adapter only has to learn a task-specific mapping on top of a representation that is already shaped by the world model.

\paragraph{Counterfactual inference.} For any unit $i$ at time $t$, individual-level treatment effects are obtained by fixing the belief $b_{i,t}$ and toggling a chosen component of $z_{i,t}$ between $0$ and $1$:
\begin{equation}
\widehat{\mathrm{CATE}}_{k}(i,t; z_j) = g_k\bigl([b_{i,t}; z_{i,t}^{(j\!=\!1)}]\bigr) - g_k\bigl([b_{i,t}; z_{i,t}^{(j\!=\!0)}]\bigr),
\end{equation}
where $z^{(j=v)}_{i,t}$ denotes the action vector with component $j$ set to $v$. Because the belief is computed from features that exclude the current action, this contrast is a counterfactual in the $do$-operator sense \cite{causality}: the consumer state representation is held fixed across the two interventions.

\section{Experiments}\label{sec:exp}

We evaluate the proposed model on a controlled simulation in which each consumer's latent traits are known by construction, allowing direct comparison between estimated CATE and the underlying ground-truth heterogeneity.

\subsection{Simulation Setup}\label{sec:sim}

Each synthetic consumer $i$ is endowed with three latent parameters that govern their behavior:
\begin{itemize}[leftmargin=1.2em,itemsep=0pt]
\item $\alpha_i$: price sensitivity, decreasing in income and age,
\item $\gamma_i$: promotion responsiveness, increasing in loyalty,
\item $\beta_i$: base preference, increasing in income, age, and loyalty.
\end{itemize}
Through their shared dependence on demographics, $\alpha_i$ and $\beta_i$ are negatively confounded via income and age, while $\gamma_i$ and $\beta_i$ are positively confounded via loyalty; $\alpha_i$ and $\gamma_i$ have no direct confounding path. The visit utility depends on $\beta_i$ and on $\gamma_i$ multiplied by the active promotional channels; the purchase utility depends on $\beta_i$ and on $-\alpha_i \cdot \mathrm{price}$, where price is synthesized causally from the action variables ($\mathrm{price} = 100 - 5\cdot\texttt{sale1} - 3\cdot\texttt{sale2} - 5\cdot\texttt{campaign} - r_i\cdot\texttt{coupon}$). Visit and purchase outcomes are then drawn independently from a logistic-Gumbel response with noise scale $0.10$.

We simulate $N=1{,}024$ consumers over $T=365$ days, yielding $312{,}320$ training samples after train/validation/test splitting at the consumer level (validation and test sets contain $30{,}720$ samples each). The visible-layer dimension is $D_v=72$ with lag window $ws=4$, and the action vector has $D_z=5$ components. The DBM uses hidden dimensions $64$--$32$--$16$ (belief dimension $112$), each adapter is an MLP with hidden dimensions $64$--$32$--$16$ trained for up to $100$ epochs with patience $30$, and the baseline MLP shares the adapter architecture but consumes raw features instead of the belief.

\subsection{Energy-Based Consistency Evaluation}\label{sec:exp-energy}

\begin{figure*}[t]
\centering
\includegraphics[width=\textwidth]{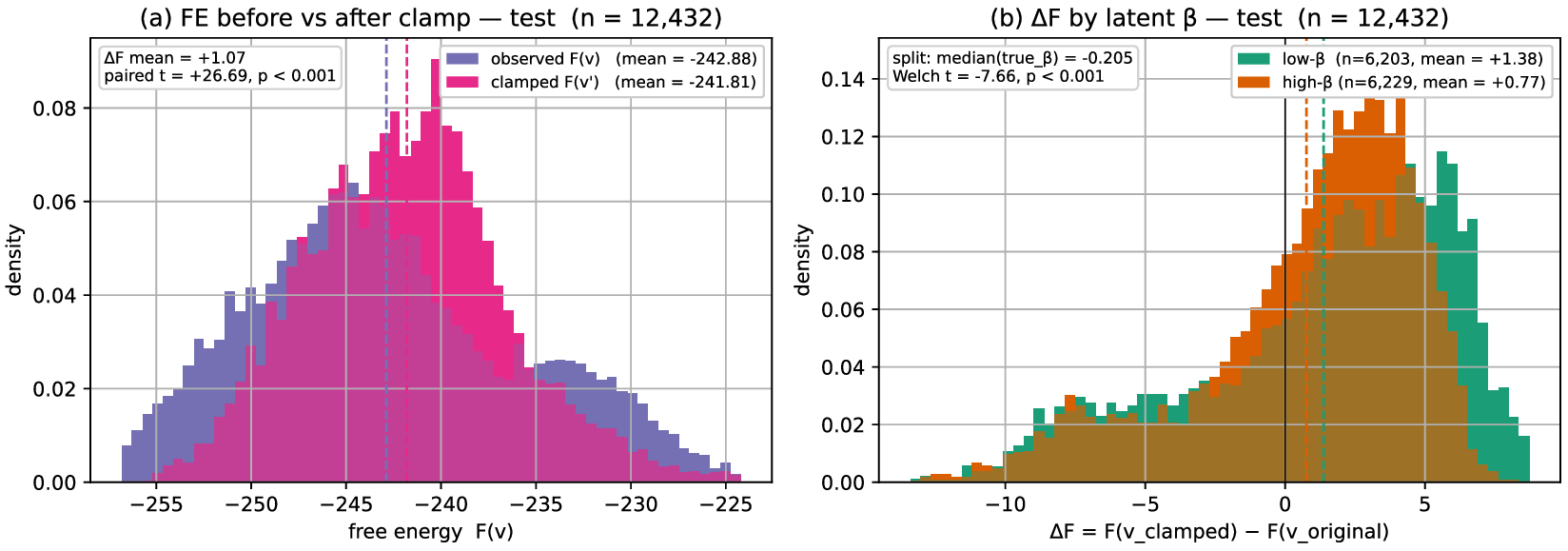}
\caption{\textbf{Energy-based consistency under the ``purchasing without recent promotion'' clamp} (test split, eligible $n=12{,}432$). \textbf{(a)} Free energy distributions before (\emph{observed $F(v)$}) and after (\emph{clamped $F(v')$}) the clamp; the clamp shifts the mean rightward by $\overline{\Delta F}=+1.07$. \textbf{(b)} Distributions of $\Delta F = F(v_\text{clamped}) - F(v_\text{original})$ split at the median of the ground-truth base preference $\beta_i$; the high-$\beta$ group (\emph{orange}) receives a uniformly smaller penalty than the low-$\beta$ group (\emph{teal}), although within-group dispersion remains substantial. All reported test statistics are highly significant ($p<0.001$); cross-split numerical values are in Table~\ref{tab:energy}.}
\label{fig:deltafe}
\end{figure*}

We probe whether the DBM's free energy in \eqref{eq:freeenergy} responds to a counterfactually clamped trajectory in a manner that reflects latent consumer heterogeneity. From each split we keep samples whose visible vector originally records no recent purchase ($\texttt{purchase\_lag}_{1{:}4}=0$), and clamp the visible layer to a ``purchasing without recent promotion'' configuration: $\texttt{purchase\_lag}_{1{:}4}=1$, $\texttt{purchased\_within\_7d}=1$, and $\texttt{campaign\_lag}_{1{:}4}=\texttt{push\_lag}_{1{:}4}=0$. All other visible units are left at their observed values. The free energy change
\begin{equation}
\Delta F(v) = F(v_\text{clamped}) - F(v_\text{original})\label{eq:deltaf}
\end{equation}
quantifies how much less plausible the DBM judges the clamped trajectory than the original. Figure~\ref{fig:deltafe} visualizes the resulting distributions on the test split, and Table~\ref{tab:energy} reports the corresponding statistics across all three splits.

\begin{table*}[t]
\centering
\caption{Energy-based consistency across data splits. Panel (a): $\Delta F$ tested against zero. Panel (b): $\Delta F$ split at the median of the ground-truth base-preference parameter $\beta_i$. Eligible $n$ counts samples whose visible vector originally has $\texttt{purchase\_lag}_{1{:}4}=0$. Within-group standard deviations of $\Delta F$ are reported in parentheses.}
\label{tab:energy}
\setlength{\tabcolsep}{6pt}
\begin{tabular}{lrrr}
\toprule
 & Train & Val & Test \\
\midrule
Eligible $n$ & $129{,}568$ & $12{,}694$ & $12{,}432$ \\
\midrule
\multicolumn{4}{@{}l}{\emph{(a) Penalty against zero}} \\
Paired $t$ & $198.25$ & $24.68$ & $26.69$ \\
$p$ (paired $t$) & ${<}0.001$ & ${<}0.001$ & ${<}0.001$ \\
$p$ (Wilcoxon) & ${<}0.001$ & ${<}0.001$ & ${<}0.001$ \\
\midrule
\multicolumn{4}{@{}l}{\emph{(b) Median split on $\beta_i$}} \\
$n_\text{high}/n_\text{low}$ & $64{,}792/64{,}776$ & $6{,}352/6{,}342$ & $6{,}229/6{,}203$ \\
$\overline{\Delta F}$ high-$\beta$ & $1.85\,(3.84)$ & $0.65\,(4.18)$ & $0.77\,(4.11)$ \\
$\overline{\Delta F}$ low-$\beta$ & $3.09\,(4.97)$ & $1.36\,(4.93)$ & $1.38\,(4.79)$ \\
Welch $t$ & $-50.27$ & $-8.64$ & $-7.66$ \\
$p$ (Welch) & ${<}0.001$ & ${<}0.001$ & ${<}0.001$ \\
$p$ (Mann--Whitney) & ${<}0.001$ & ${<}0.001$ & ${<}0.001$ \\
\bottomrule
\end{tabular}
\end{table*}

Table~\ref{tab:energy}(a) shows that $\Delta F$ is systematically positive on every split: the DBM treats ``the consumer purchased on every recent day but received neither campaigns nor pushes'' as substantially less plausible than the corresponding observed trajectory, with overwhelming significance under both a paired $t$-test and a Wilcoxon signed-rank test. Because the clamp targets lagged actions and outcomes rather than a single-period co-occurrence pattern, this is a property the static DBM of \cite{niimi_boltzmanngpt} cannot exhibit: the consistency check here operates on time-extended trajectories.

Table~\ref{tab:energy}(b) further shows that the penalty is heterogeneous in the latent base preference. Splitting the eligible samples at the median of the ground-truth $\beta_i$ yields a smaller mean $\overline{\Delta F}$ for the high-$\beta$ group than for the low-$\beta$ group on every split, with the gap significant under both Welch's $t$-test and the Mann--Whitney $U$ test. ``Purchasing without promotion'' is therefore less surprising to the DBM precisely when the consumer's observed covariates already predict a high intrinsic preference. The DBM is trained without access to $\beta_i$, so this dependence emerges purely through the demographic-to-belief mapping that the world model learns. This is the energy-side counterpart of the disentanglement signal observed in CATE recovery (Section~\ref{sec:exp-cate}): the same belief that supports counterfactual contrasts in the adapter also shapes the energy landscape in a $\beta$-aware direction.

\subsection{Prediction Performance}\label{sec:exp-pred}

We first verify that compressing the $72$-dimensional input into a $112$-dimensional frozen belief does not damage predictive power. Table~\ref{tab:auc} compares the test-set AUC of the baseline MLP, which consumes the raw features and the current action, against the adapter that consumes the frozen belief and the current action. The two are within $0.001$--$0.002$ AUC of each other on both tasks. The adapter even matches or slightly exceeds the baseline on visit prediction. We conclude that the DBM's representation retains the predictive content of the input.

\begin{table}[t]
\centering
\caption{Test-set AUC for visit and purchase prediction. \emph{Baseline MLP} consumes raw features and the current action; \emph{Belief Adapter} consumes the frozen belief and the current action.}
\label{tab:auc}
\begin{tabular}{lccc}
\toprule
Task & MLP & Belief Adapter & Gap \\
\midrule
visit & $0.7119$ & $0.7130$ & $+0.0011$ \\
purchase & $0.6777$ & $0.6772$ & $-0.0005$ \\
\bottomrule
\end{tabular}
\end{table}

\subsection{CATE Recovery}\label{sec:exp-cate}

We next evaluate whether the adapter, when used for counterfactual prediction, recovers the true heterogeneity of treatment effects across consumers. We consider two interventions whose causal structure differs:
\begin{itemize}[leftmargin=1.2em,itemsep=0pt]
\item \texttt{push}\,$\to$\,visit, whose effect in the data-generating process is governed by the promotion-responsiveness parameter $\gamma_i$ and which is unconfounded with $\alpha_i$;
\item \texttt{sale1}\,$\to$\,purchase, whose effect operates through the price channel and is governed by the price-sensitivity parameter $\alpha_i$, with $\alpha_i$ and $\beta_i$ confounded through income and age.
\end{itemize}
For each intervention, we compute estimated CATE on the test set under six models---S-learner, T-learner, X-learner, DR-learner, Causal Forest \cite{causalforest,metalearners,drlearner}, and the proposed Belief Adapter---all using the same raw features for the meta-learners and the frozen belief for the adapter. The X-learner and DR-learner use random-forest base learners, with the DR-learner additionally relying on $2$-fold cross-fitting for its nuisance estimates. As the principal metric, we report the Spearman rank correlation $\rho$ between the estimated CATE and each true latent parameter on the logit scale, which removes the sigmoid compression that distorts probability-scale CATE.

\begin{table*}[t]
\centering
\caption{CATE recovery: Spearman $\rho$ between estimated CATE (logit scale) and true latent parameters on the test set. \textbf{Bold} indicates the best recovery of the target parameter for each intervention. \emph{S}: S-learner. \emph{T}: T-learner. \emph{XL}: X-learner with random-forest base learners \cite{metalearners}. \emph{DR}: DR-learner \cite{drlearner} with random-forest nuisance models and 2-fold cross-fitting. \emph{CF}: Causal Forest. \emph{Adapter}: belief adapter (proposed). XL, DR, and CF report risk-difference uplift, so the logit-scale and probability-scale columns coincide for these estimators.}
\label{tab:cate}
\setlength{\tabcolsep}{4pt}
\begin{tabular}{llcccccc}
\toprule
Intervention & Param & S & T & XL & DR & CF & Adapter \\
\midrule
\multirow{3}{*}{\texttt{push}\,$\to$\,visit}
 & $\alpha$ (irrelevant) & $+0.004$ & $-0.284$ & $+0.043$ & $-0.008$ & $+0.071$ & $-0.148$ \\
 & $\beta$ (confound)    & $+0.204$ & $+0.441$ & $+0.067$ & $+0.079$ & $+0.063$ & $+0.377$ \\
 & $\gamma$ (target)     & $+0.579$ & $+0.558$ & $+0.318$ & $+0.194$ & $+0.350$ & $\mathbf{+0.625}$ \\
\midrule
\multirow{3}{*}{\texttt{sale1}\,$\to$\,purchase}
 & $\alpha$ (target)    & $+0.406$ & $+0.161$ & $+0.123$ & $-0.016$ & $+0.040$ & $\mathbf{+0.548}$ \\
 & $\beta$ (confound)   & $-0.394$ & $-0.154$ & $-0.154$ & $-0.004$ & $-0.044$ & $-0.597$ \\
 & $\gamma$ (irrelevant)& $-0.043$ & $-0.027$ & $-0.060$ & $-0.042$ & $-0.006$ & $-0.306$ \\
\bottomrule
\end{tabular}
\end{table*}

Table~\ref{tab:cate} summarizes the results. On both interventions, the adapter attains the highest rank correlation between estimated CATE and the target latent parameter. For the unconfounded \texttt{push}\,$\to$\,visit intervention, the gap is moderate ($\rho_{\gamma}=+0.625$ vs.\ $+0.579$ for the strongest meta-learner baseline). For the confounded \texttt{sale1}\,$\to$\,purchase intervention, the gap widens substantially: the adapter recovers $\rho_{\alpha}=+0.548$, whereas the T-learner and Causal Forest, which are well known to be sensitive to limited overlap and confounded treatment assignment \cite{metalearners,causalforest}, drop to $+0.161$ and $+0.040$, respectively. The X-learner and DR-learner, despite being explicitly designed to mitigate the limitations of T-learner and to deliver doubly robust pseudo-outcome regression, only reach $\rho_{\alpha}=+0.123$ and $-0.016$ in this setting, plausibly because their random-forest base learners and cross-fitting introduce additional variance on top of an already noisy treated/control split. The S-learner remains competitive at $+0.406$ but is still outperformed by the adapter by more than $0.14$ in $\rho$.

A second observation concerns leakage to nuisance parameters. Although the X-learner, DR-learner, and Causal Forest record relatively low absolute correlations with the target, they also exhibit small magnitudes on the irrelevant parameters: on \texttt{push}\,$\to$\,visit, $|\rho_\alpha|$ stays below $0.08$ for all three, and on \texttt{sale1}\,$\to$\,purchase, $|\rho_\gamma|$ stays below $0.07$. The S-learner and T-learner show larger spurious correlations with the confound parameter, reflecting their reliance on a single outcome model whose treatment-feature interactions can absorb confounded variation. The adapter, by contrast, achieves both the strongest target-parameter recovery and a sizable confound correlation that, by construction of the simulator, is in the direction implied by the latent confounding path.

The adapter also exhibits stronger negative correlation with the confound ($\beta$) on \texttt{sale1}\,$\to$\,purchase ($\rho_{\beta}=-0.597$). Because $\alpha_i$ and $\beta_i$ are negatively confounded by construction, this is consistent with the adapter relying on $\alpha$-aligned information rather than on $\beta$-aligned shortcuts; we caution, however, that with a single confounding path the two interpretations cannot be fully distinguished from this experiment alone. The cleaner \texttt{push}\,$\to$\,visit result, in which $\gamma$ is recovered while $\alpha$ remains near zero, supports the view that the belief carries identifiable, separable information about distinct latent traits.

Taken together, Table~\ref{tab:auc} and Table~\ref{tab:cate} indicate that the DBM's belief representation does not merely match raw-feature predictors on outcome accuracy but provides a substrate on which counterfactual contrasts recover heterogeneous effects more faithfully than standard meta-learners (S-, T-, X-, and DR-learner) and Causal Forest applied to the same features.

\section{Conclusion}\label{sec:conc}

We have proposed a Three-in-One world-model architecture for marketing intervention in which a frozen DBM provides a belief representation that simultaneously supports energy-based consistency evaluation, outcome prediction, and counterfactual inference. The design separates a heavyweight world model from lightweight task-specific adapters, mirroring the asymmetry between environment models and controllers in \cite{ha2018}, and incorporates time-varying information through lagged features in the visible layer instead of recurrent connections.

In a controlled simulation with known latent traits, prediction adapters matched a strong MLP baseline on test-set AUC for visit and purchase while recovering heterogeneous treatment effects substantially better than S-, T-, X-, and DR-learner meta-learners and a Causal Forest baseline. The advantage was largest for a confounded price-promotion intervention, where the price-sensitivity parameter is easily masked by the negatively confounded base preference. Free-energy clamps further showed that the DBM penalizes counterfactual purchase trajectories without prior promotional exposure, with the penalty itself heterogeneous in the latent base preference---an energy-side counterpart to the disentanglement signal observed in CATE recovery. These findings suggest that DBM-based world models can provide a unified substrate for prediction, consistency evaluation, and causal-effect estimation in marketing settings where latent heterogeneity drives intervention response.

Several limitations remain. All experiments rely on a single simulation seed and a single data-generating process; replication across seeds and across alternative DGPs is needed to characterize the variance and the conditions under which the advantages of the world-model adapter and of the energy-based consistency signal persist. The energy-based analysis additionally examines a single clamp specification (\emph{purchasing without recent promotion}) and reports a population-level effect: although the high-versus-low-$\beta$ gap in $\overline{\Delta F}$ is highly significant, within-group dispersion is large enough that $\Delta F$ alone is not a sample-level classifier of latent type. Sensitivity to alternative counterfactual configurations---for example, implausible promotional bursts on consumers whose history shows no engagement---remains to be characterized. We also restrict attention to a single product category and binary outcomes; extension to multiple products via choice models, to continuous outcomes such as session length and basket value, and to real A/B-test data are natural next steps. Finally, learning optimal intervention policies on top of the world model---closing the loop between belief, adapter, and controller in the sense of \cite{ha2018}---remains for future work.

\bibliographystyle{unsrt}
\bibliography{purchaseworld}

\end{document}